\documentclass[runningheads]{llncs}

\usepackage{eccv}

\usepackage{eccvabbrv}

\usepackage{graphicx,tcolorbox,subfiles}
\usepackage{booktabs}
\usepackage{color, colortbl}
\definecolor{Gray}{gray}{0.9}
\definecolor{DarkGray}{gray}{0.7}

\usepackage[accsupp]{axessibility}  

\usepackage{hyperref}

\usepackage{orcidlink}

\begin{document}

\title{An Augmentation-based Model Re-adaptation Framework for Robust Image Segmentation}


\titlerunning{Augmentation-based Model Re-adaptation Framework}

\author{Zheming Zuo\inst{1}\orcidlink{0000-0003-1576-0865} \and
Joseph Smith\inst{1} \and \\Jonathan Stonehouse\inst{2} \and
Boguslaw Obara\inst{1}\orcidlink{0000-0003-4084-7778}}

\authorrunning{Z.~Zuo et al.}

\institute{School of Computing, Newcastle University, UK\\
\email{\{zheming.zuo,j.smith57,boguslaw.obara\}@newcastle.ac.uk}
\and
Procter and Gamble, UK \hspace{0.5em}
\email{stonehouse.jr@pg.com}
}

\maketitle

\begin{abstract}

Image segmentation is a crucial task in computer vision, with wide-ranging applications in industry. The Segment Anything Model (SAM) has recently attracted intensive attention; however, its application in industrial inspection, particularly for segmenting commercial anti-counterfeit codes, remains challenging. Unlike open-source datasets, industrial settings often face issues such as small sample sizes and complex textures. Additionally, computational cost is a key concern due to the varying number of trainable parameters. To address these challenges, we propose an Augmentation-based Model Re-adaptation Framework (AMRF). This framework leverages data augmentation techniques during training to enhance the generalisation of segmentation models, allowing them to adapt to newly released datasets with temporal disparity. By observing segmentation masks from conventional models (FCN and U-Net) and a pre-trained SAM model, we determine a minimal augmentation set that optimally balances training efficiency and model performance. Our results demonstrate that the fine-tuned FCN surpasses its baseline by 3.29\% and 3.02\% in cropping accuracy, and 5.27\% and 4.04\% in classification accuracy on two temporally continuous datasets. Similarly, the fine-tuned U-Net improves upon its baseline by 7.34\% and 4.94\% in cropping, and 8.02\% and 5.52\% in classification. Both models outperform the top-performing SAM models (ViT-Large and ViT-Base) by an average of 11.75\% and 9.01\% in cropping accuracy, and 2.93\% and 4.83\% in classification accuracy, respectively.


\end{abstract}

\section{Introduction}
\label{sec:intro}

\begin{figure}[tb]
  \centering
  \includegraphics[width=.97\linewidth]{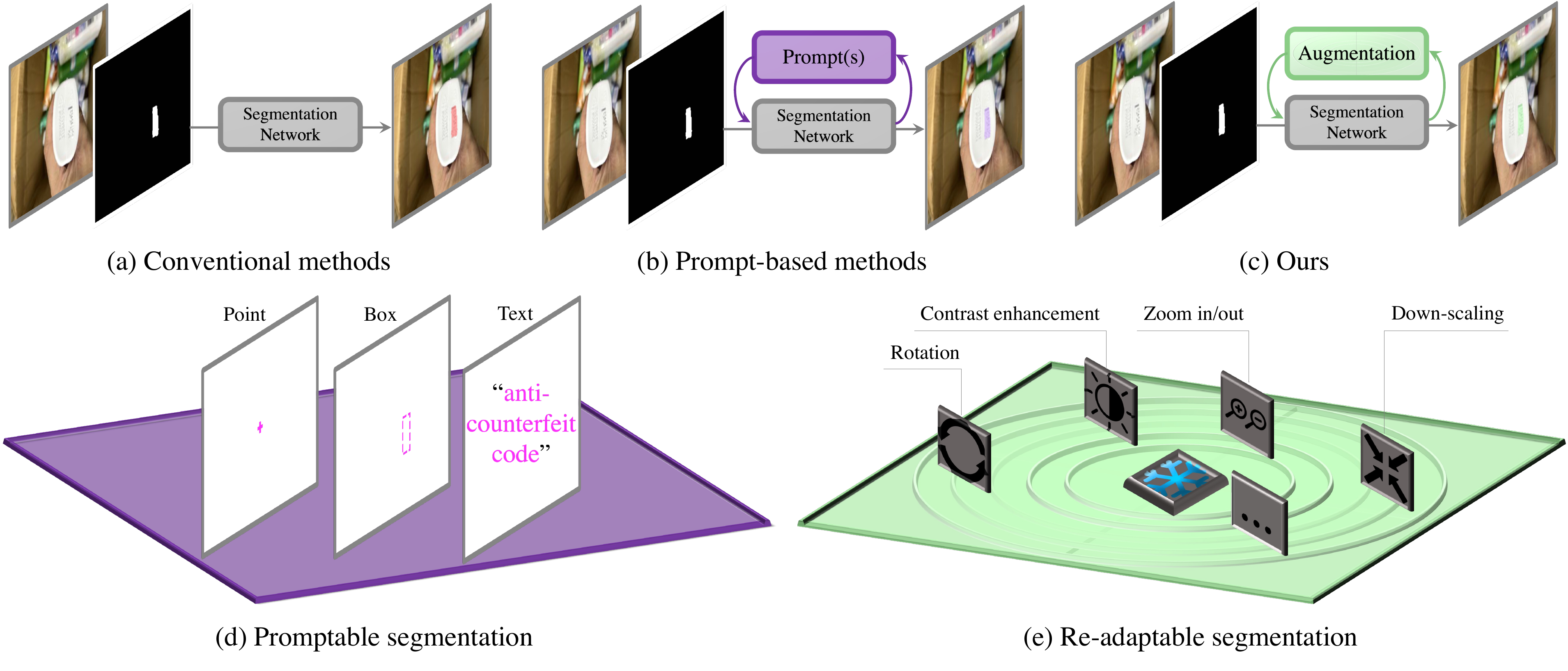}
  \caption{Different from segmentation workflows of (a) straightforward yet less-informed \emph{e.g.} FCN \cite{long2015fully} and U-Net \cite{ronneberger2015u}, as well as (b) interactive and more-informed \emph{e.g.} FocalClick \cite{chen2022focalclick}, PseudoClick \cite{liu2022pseudoclick} and SAM \cite{kirillov2023segment} using prompts (d), our augmentation-based model re-adaptation framework (c), embraces the concept of pseudo re-adaptation (Sec. \ref{sec:pr}) by interacting with the currently trained model as depicted in (e), minimising the number of candidates in the augmentation pool to be adopted in the training phase with only the earliest released dataset, while maximising the accuracies of both segmentation and classification accuracy.}
  
  \label{fig:mot}
\end{figure}

Deep learning has revolutionised the field of computer vision \cite{minaee2021image,wei2024physical,mettes2024hyperbolic} in the past decade, enabling more precise and efficient visual data analysis. Image segmentation, the process of partitioning an image into meaningful regions, is crucial for various applications, including brain tumour lesions detection \cite{siddique2021u}, obstacle identification \cite{li2023mseg3d,shoeb2024have}, and facial recognition \cite{chen2022harnessing,li2023robust}. Deep neural networks, particularly Convolutional Neural Networks (CNNs) \cite{lecun1998gradient}, depicted in Fig. \ref{fig:mot}(a), have 

\noindent significantly advanced the capabilities of image segmentation by learning complex features from large-scale datasets. Representative models \emph{e.g.} Fully Convolutional Networks (FCNs) \cite{long2015fully}, U-Net \cite{ronneberger2015u}, and Mask Region-based Convolutional Neural Networks (R-CNN) \cite{he2017mask} have been commonly employed. FCNs replace traditional fully connected layers with convolutional layers to produce segmentation maps. U-Net, known for its encoder-decoder architecture, excels in biomedical image segmentation. Meanwhile, anchor-based methods \emph{e.g.} Mask R-CNN extend Faster R-CNN \cite{ren2015faster} to improve object detection. These conventional models exemplify the power of deep learning in achieving competitive performance in image segmentation tasks.

Notably, segmentation is increasingly in demand for commercial product images, particularly in brand protection and inventory management \cite{wang2022comprehensive}. Interested regions \emph{e.g.} logos \cite{hou2023deep}, labels \cite{jia2019coarse}, or barcodes \cite{ni2020effective}, facilitating further analysis including cropping and classification. Recently, SAM \cite{kirillov2023segment}, as a versatile approach, can precisely segment visual objects without any additional training (\emph{i.e.} prompt-based zero-shot mask prediction in Fig. \ref{fig:mot}(b)), making it highly effective for applications in retail and consumer goods where it can identify and segment household items or product packaging \cite{huang2024segment}. In brand protection, image segmentation is crucial in systems that separate interested regions from an image, \emph{e.g.} printed dot-matrix codes on commercial products \cite{jia2019coarse}. This segmentation is essential for enabling further classification processes in the segmented areas. The competitive method employs CNNs to segment dot-matrix codes from commercial product images efficiently. However, the performance of these models tends to deteriorate under sub-optimal conditions as well as evolving changes and bias in code patterns along with the chronological line \cite{jackson2021camera}.


Typical challenges associated with segmentation tasks include varying lighting conditions \emph{e.g.} low-light \cite{yang2020advancing} or over-exposure \cite{fu2023raw}, different degrees of blurriness caused by unstable camera motion \cite{Zuo_IdeaNet_2022_WACV}, distortion \cite{li2023fg} or stretch \cite{zhang2024behind} in the captured images, which in turn demand higher computation cost.
Practically, suppose that $\mathcal{D}_{t_1}$, $\mathcal{D}_{t_2}$, and $\mathcal{D}_{t_3}$ are the three continuous datasets with a certain degree of disparity along the time points, $t_1, t_2$ and $t_3$, and a baseline model has been trained on $\mathcal{D}_{t_1}$; this work aims to find a way of fine-tuning the model trained on $\mathcal{D}_{t_1}$ to maximise the cropping and classification accuracies on $\mathcal{D}_{t_2}$ and $\mathcal{D}_{t_3}$ through a set of augmentation methods which is minimised, illustrated in Fig. \ref{fig:mot}(c). Noteworthy, the Ground Truth (GT) masks of $\mathcal{D}_{t_1}$ are available, whereas the others are not. Such an optimisation process can be expressed by:

\begin{equation}\label{dq:opt}
\operatorname*{argmax}_{\theta \sim \mathcal{A}(\mathcal{D}_{t_1})} \left( \text{Acc}\big(\mathcal{M}_{\text{crop}}^{\theta}(\mathcal{D}_{t_2}, \mathcal{D}_{t_3})\big), \text{Acc}\big(\mathcal{M}_{\text{cla}}^{\phi}(\mathcal{D}_{t_2}, \mathcal{D}_{t_3})\big) \right) \, \text{s.t.}  \, \operatorname*{argmin}_{\mathcal{A}} |\mathcal{A}|,
\end{equation}

where $\theta$ represents the weights optimised for the segmentation model $\mathcal{M}_{\text{crop}}$, which is responsible for cropping images in datasets $\mathcal{D}_{t_2}$ and $\mathcal{D}_{t_3}$. $\phi$ denotes the pre-trained weights of the image classifier $\mathcal{M}_{\text{cla}}$ used for classifying the cropped images from $\mathcal{D}_{t_2}$ and $\mathcal{D}_{t_3}$. $|\mathcal{A}|$ represents the size of the augmentation pool $\mathcal{A}$. In practice, the performance of both cropping and its associated classification could be enhanced by introducing several off-the-shelf data augmentation methods during the training phase; we expect to conclude a minimum set of augmentation operations as an effort to contribute to reducing the space and time complexity of model training for automatic industrial inspection.

Fundamentally, the choice of augmentation methods highly relies on the data scale and pattern disparity between datasets $\mathcal{D}_{t_1}$,  $\mathcal{D}_{t_2}$ and $\mathcal{D}_{t_3}$. However, those characteristics cannot be precisely quantified as the existence of challenging factors mentioned before the antecedent paragraph and various types of uncertainties (\emph{e.g.}~dataset noise and model randomness) \cite{jungmann2024analytical} raised during the training phase.  In the meantime, the large-scale pertaining may not be sufficiently capable of cropping our work's Anti-CounterFeit (ACF) code.

This work aims to fill in the gap of robust image segmentation with a minimum cost of training or fine-tuning the segmentation network from a different yet explainable perspective. Generally, we present an automatic framework for determining the smallest amount of augmentation methods involved in training, dubbed Augmentation-based Model Re-adaptation Framework (AMRF), for an image segmentation task in the use of industrial inspection given three chronologically continued datasets and the earliest one is used for training merely, and our contributions are summarised as follows:

\textbf{1)} We propose an automatic augmentation method determination framework for training (Sec.~\ref{sec:eap}) through the components of adaptive-angle cropping (Sec. \ref{sec:angle}) and the pseudo re-adaption configured in a curriculum learning manner.

\textbf{2)} We prove the feasibility and efficacy of the proposed framework on two conventional segmentation networks in surpassing the pre-trained SAM even by providing box prompts with a large margin concerning cropping (as an intermediate segmentation indication in Sec. \ref{sec:pm}) and classification precision.

\textbf{3)} We show that performing pseudo re-adaption (Sec. \ref{sec:pr}) could be a feasible starting point for fine-tuning a light-weight segmentation network for deploying the proposed framework in diverse industrial scenarios, where merely the model with the smallest amount of weights learned from the earliest data release (Sec. \ref{sec:ds}) is required to manage temporal knowledge drift.

\section{Related Work and Dataset}


\subsection{Deep Segmentation Networks}

\subsubsection{Conventional Segmentation Networks.}

CNNs have become the cornerstone of modern image segmentation, with several architectures achieving state-of-the-art performance. The U-Net architecture \cite{ronneberger2015u}, initially designed for biomedical image segmentation, employs an encoder-decoder structure with skip connections, enabling precise localisation and context utilisation. FCNs \cite{long2015fully} introduced end-to-end training for segmentation tasks, converting fully connected layers into convolutional ones to output spatially consistent predictions. Mask R-CNN \cite{he2017mask} extends Faster R-CNN \cite{ren2015faster} by adding a parallel branch for predicting segmentation masks, effectively handling instance segmentation. Pyramid Scene Parsing Network (PSPNet) \cite{zhao2017pspnet} incorporates pyramid pooling to capture contextual information at multiple scales, significantly enhancing the segmentation accuracy in complex scenes. These methods collectively demonstrate the robustness and versatility of CNNs in addressing diverse segmentation challenges, forming the foundation for many subsequent advancements in this field.

\subsubsection{Transformer-based Segmentation Networks.}

Recently, transformer-based models in conjunction with prompt-based approaches have revolutionised image segmentation by leveraging attention mechanisms and contextual embeddings \cite{han2023transformer, liu2023survey, thisanke2023semantic}. SAM \cite{kirillov2023segment} exemplifies this trend by employing transformers to predict high-quality segmentation masks from prompts, significantly improving flexibility and accuracy. FocalClick \cite{chen2022focalclick} advances this by enabling practical, interactive image segmentation with a focus on user-friendly prompt designs. PseudoClick \cite{liu2022pseudoclick} leverages prompt-based methods to enhance the segmentation performance by iteratively refining pseudo-labels, demonstrating robustness in scenarios with limited labelled data. OneFormer \cite{jain2023oneformer} employs a unified transformer-based architecture to perform various image segmentation tasks by learning to interpret and segment images with high precision, leveraging its ability to understand contextual relationships within the image data. Models utilising Contrastive Language-Image Pre-training(CLIP) \cite{radford2021learning} bridge the gap between vision and language, enabling zero-shot segmentation through the alignment of visual and textual representations \cite{luddecke2022image}. These innovations underscore the potential of transformers and prompt-based methods in advancing the state of the art, offering new paradigms for image segmentation that excel in precision and generalisation capabilities across diverse datasets.

\subsubsection{Importance of Image Quality in Segmentation and the Role of Augmentation.}

Image quality is closely related to the performance of segmentation networks, as higher quality images typically contain more discernible features and details essential for accurate segmentation \cite{yu2023techniques}. Variations in lighting, noise, resolution, and occlusions can significantly impact the effectiveness of segmentation algorithms. Data augmentation techniques are widely employed to mitigate these challenges and improve the robustness of segmentation models \cite{shorten2019survey}. Augmentation strategies such as rotation, scaling, cropping, flipping, and adding noise help create diverse training datasets, allowing models to generalise better to different scenarios and enhancing their resilience to variations in real-world data. By artificially expanding the training dataset, augmentation helps prevent overfitting. Also, it ensures that the model can handle a wide range of input conditions, ultimately leading to more reliable and accurate segmentation outcomes. This emphasis on quality and diversity in training data is equally critical in image classification tasks, where the goal is to assign labels to images based on their visual content accurately \cite{img_quality_2024,Zuo_DCF_2024_CVPRW}.

\subsubsection{Existing Augmentation Methods for Segmentation.}

Existing augmentation methods for image segmentation encompass a variety of techniques to enhance model robustness by increasing training data diversity. Traditional methods include geometric transformations (\emph{e.g.} rotation, scaling, translation, flipping), colour adjustments (\emph{e.g.} brightness, contrast \cite{wang2021exploring}, saturation), and elastic deformations, particularly useful in medical imaging \cite{ronneberger2015u, shorten2019survey}. Noise addition (\emph{e.g.} Gaussian, salt-and-pepper, speckle) helps models handle noisy data \cite{ciresan2012deep}. Advanced techniques, such as cutout and mixup, mask or mix image parts to prevent overfitting \cite{devries2017improved, zhang2020does}. Generative Adversarial Networks (GANs) and neural style transfer generate synthetic images, enriching training datasets with realistic examples \cite{antoniou2017data}. These strategies collectively improve segmentation model accuracy and robustness across diverse conditions and applications.

\subsection{Datasets and Challenging Factors} \label{sec:ds}

\begin{figure}[tb]
  \centering
  \includegraphics[width=.97\linewidth]{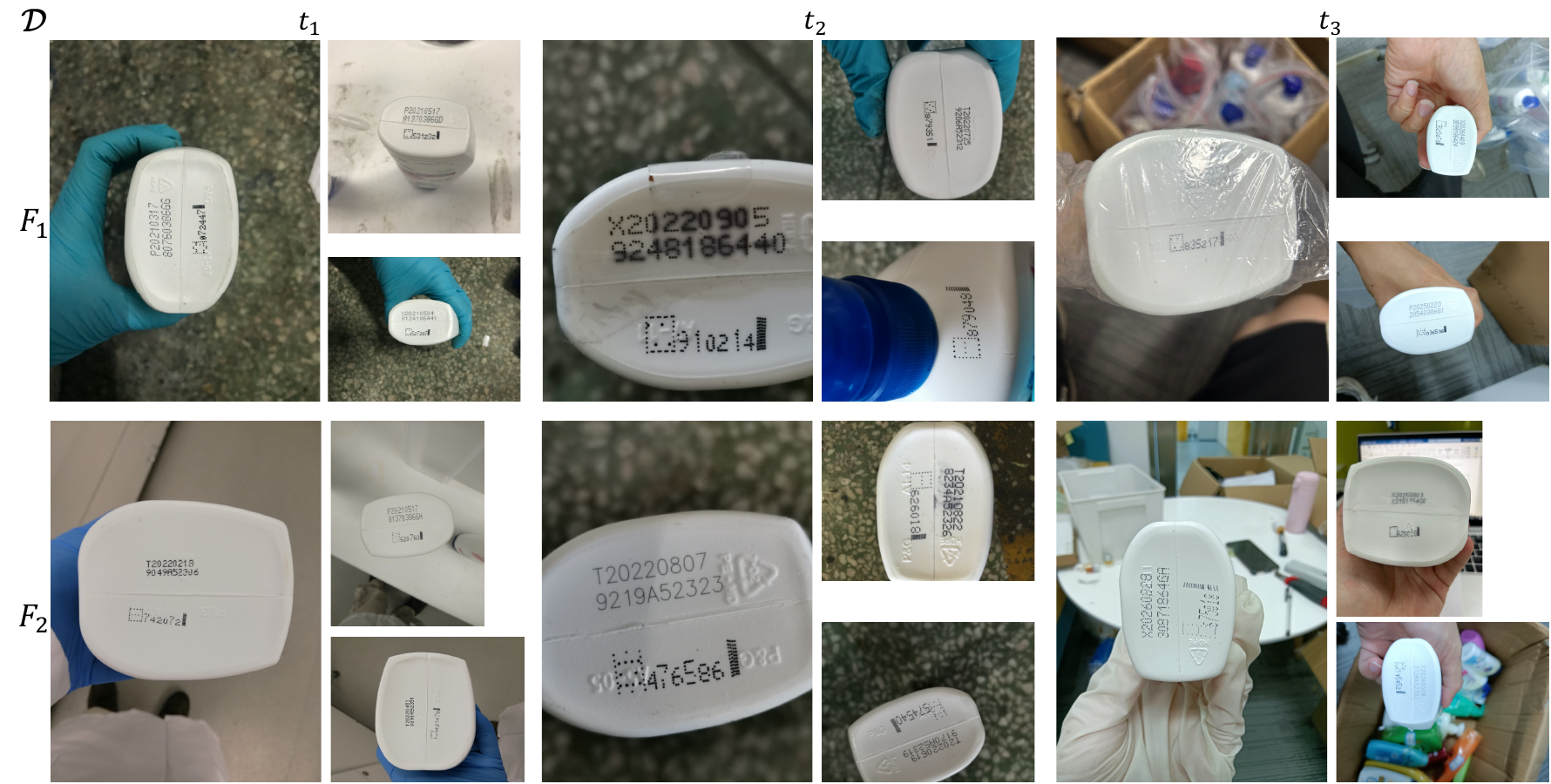}
  \caption{Sample images of the bottom side of commercial products produced by two factories, $F_1$ and $F_2$, in three temporally continuous datasets, \emph{i.e.} $\mathcal{D}_{t_1}$, $\mathcal{D}_{t_2}$ and $\mathcal{D}_{t_3}$. The ACF code is the region of interest required to be segmented and cropped.
  }
  \label{fig:ds}
\end{figure}

To facilitate the conduct of this research, our industrial partners delivered three RGB datasets that are continuous in the timeline, $\mathcal{D}_{t_1}$, $\mathcal{D}_{t_2}$ and $\mathcal{D}_{t_3}$ of mostly the bottom side of commercial products produced by two factories $F_1$ and $F_2$, partially presented in Fig.~\ref{fig:ds}.

Datasets $\mathcal{D}_{t_1}$, $\mathcal{D}_{t_2}$ and $\mathcal{D}_{t_3}$ contain 1370, 1716 and 1093 RGB images, each of which comes with different resolutions due to evolving types of cameras were adopted for data-capturing in uncontrolled scenes. This results in the degree of disparity among these three datasets increasing along the chronological line, accordingly making the ACF code segmentation challenging. The code starts with a box of dots followed by different-sized numbers and then lines of dots.

Noteworthy, $\mathcal{D}_{t_1}$ is the only dataset with images and masks in this work. In this work, we aim to figure out a way to enhance the model generalisation capability in automatically segmenting ACF code by training on $\mathcal{D}_{t_1}$ merely and testing on the entire $\mathcal{D}_{t_2}$ and $\mathcal{D}_{t_3}$. Concretely, $\mathcal{D}_{t_1}$ is divided into a training set and a validation set in a ratio of 4:1. Additionally, as an effort to contribute to quantitative analysis of industrial inspection, the distribution of samples associated with factories $F_1$ and $F_2$ are as follows: $|\mathcal{D}_{t_1}^{F_{1}}|$ = 965 and  $|\mathcal{D}_{t_1}^{F_{2}}|$ = 405, $|\mathcal{D}_{t_2}^{F_{1}}|$ = 1389 and  $|\mathcal{D}_{t_2}^{F_{2}}|$ = 327, $|\mathcal{D}_{t_3}^{F_{1}}|$ = 375 and  $|\mathcal{D}_{t_3}^{F_{2}}|$ = 718.

Several issues must be considered to address the practical challenges of the three datasets. Firstly, the varying distances between the camera and the bottom side of the product can result in images that are either too close or too far, affecting the segmentation accuracy. Secondly, image blur caused by lens shake or lack of focus can further complicate the segmentation process. Thirdly, uneven surface textures, particularly those extending from the printed ACF area, pose additional challenges. Furthermore, the last vertical stripe-shaped dot part of the ACF code may sometimes be far away or not aligned horizontally with the rest of the code, complicating the segmentation task. Lastly, surface stains, irrelevant stickers, plastic films, and splashes left over from printing can obscure the region of interest, making accurate segmentation more difficult.

\section{Method}\label{sec:met}



\subsection{Workflow}

\begin{figure}[tb]
  \centering
  \includegraphics[width=.98\linewidth]{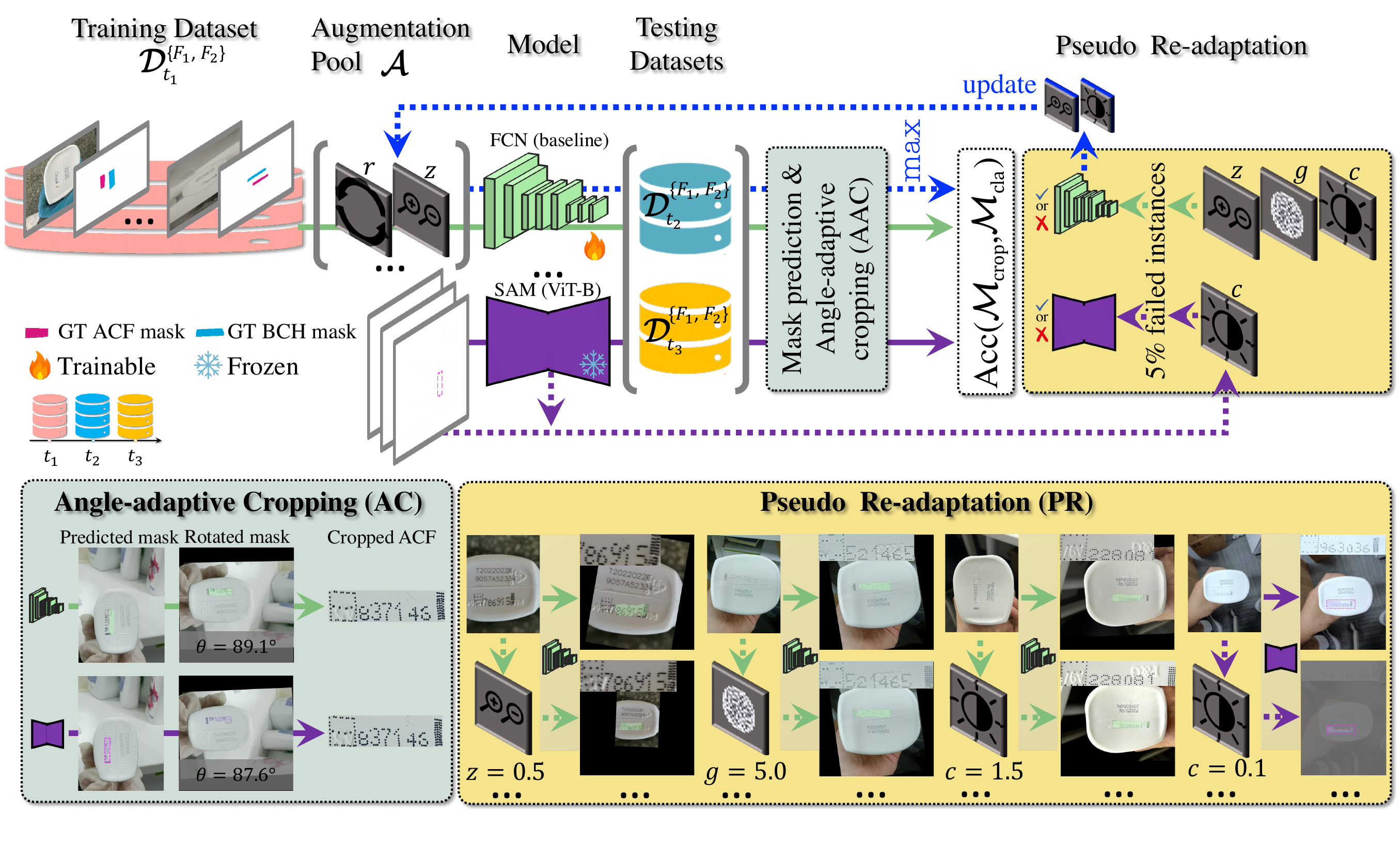}
  \caption{Workflow of the proposed AMRF for robust image segmentation in industrial inspection. The figure highlights the enhanced augmentation pool, where candidate augmentation methods are included based on their ability to improve cropping and classification accuracies over the current baseline. Cropping accuracy is evaluated by segmenting the region of interest and adopting an angle-adaptive cropping component to align the ACF code horizontally (described in Sec.~\ref{sec:angle}). The flame symbol highlights the trainable FCN model (trained using $\mathcal{D}_{t_1}^{\{F_{1},F_{2}\}}$ and the given GT binary masks), whereas the snowflake symbol denotes the SAM model with frozen weights. In contrast, classification accuracy is assessed using an existing code classifier. A key component of AMRF, pseudo re-adaptation, modifies the input image to align with the current model's perception range to inform the gradual expansion of the augmentation pool. $z$, $g$, $c$ and $r$ refer to the variable settings for the augmentation methods.
  }
  \label{fig:framework}
\end{figure}

The Augmentation-based Model Re-adaptation Framework (AMRF), as the workflow depicted in Fig.~\ref{fig:framework}, aims to improve the image segmentation model's robustness and generalisation capabilities in industrial applications, especially when dealing with small sample sizes and challenging textures that partially exampled in Fig. \ref{fig:ds}. The process iteratively refines the augmentation pool and re-trains the model to achieve optimal cropping performance.

The initial phase, as in the top row of Fig.~\ref{fig:framework}, involves training the baseline segmentation model using the dataset $\mathcal{D}_{t_1}$ with the default augmentation methods in the existing pool applied shown in Sec. \ref{sec:eap}. Angle-adaptive Cropping (AC), as seen in the bottom left of Fig.~\ref{fig:framework} and Sec. \ref{sec:angle}, is particularly devised to crop the ACF code region from the input image appropriately in line with the predicted mask. Post-training, the model is tested on datasets $\mathcal{D}_{t_2}$ and $\mathcal{D}_{t_3}$ to identify images where the model's performance is lacking. From this testing phase, 5\% of the images the model failed to accurately segment are selected for further analysis. These selected images are then utilised for Pseudo re-adaptation, where they undergo augmentation using candidate methods not yet included in the existing pool (refer to the bottom right of Fig.~\ref{fig:framework} and Sec. \ref{sec:pr}). The currently trained segmentation model and the pre-trained SAM jointly assess these arising candidate methods to evaluate their suitability in boosting segmentation and associated cropping accuracy.

Qualified candidate methods are then added to the augmentation pool with their determined value ranges, ensuring only complementary augmentations are included to maintain minimal complexity (blue arrow at the top of Fig.~\ref{fig:framework} and Sec. \ref{sec:eap}). The process iterates with re-training the segmentation model using the updated augmentation pool. This cycle continues until the optimal augmentation pool is determined, characterised by the smallest number of augmentation methods required to sustain or enhance model performance.

\subsection{Evolving Augmentation Pool} \label{sec:eap}

The Evolving Augmentation Pool (EAP) within our AMRF, as a solution to the optimisation problem formulated in Eq. (\ref{dq:opt}), is designed to dynamically adjust augmentation operations based on the model's predictions and specific challenges encountered during training and validation. The primary objective of EAP is to refine a pool of augmentation methods (bottom right of Fig. \ref{fig:framework}) to include the optimal methods to fine-tune a model for a dataset with multiple challenging factors. While traditional augmentation techniques like rotation and zooming are standard practices for improving model robustness, they may not sufficiently address specific issues in our datasets. The evolving nature of the augmentation pool allows us to introduce new augmentation techniques tailored to particular challenges observed during model performance evaluation.

One significant challenge EAP addresses is the variation in zoom levels caused by images captured at different scales. This discrepancy can lead to inconsistencies in model predictions. By incorporating zoom-in/out augmentations, we simulate these variations during training, enabling the model to handle real-world scale differences more effectively. Another critical challenge is the contrast differences resulting from uneven textures in the commercial products, such as varying lighting conditions and surface finishes. These variations can impact segmentation accuracy. Adjusting the contrast helps the model become more resilient to such inconsistencies. 

Additionally, Gaussian blur addresses blurriness caused by camera motion and focus issues. This augmentation helps the model learn to identify and precisely segment objects even when the images are not perfectly sharp. It also helps alleviate some of the issues with uneven textures, as the contrast is less pronounced when the image is slightly blurred.

Continuously evolving the augmentation pool ensures the model remains robust and adaptable to new challenges. This ultimately leads to improved segmentation accuracy and generalisation across diverse datasets. This dynamic approach allows us to tailor the training process to the application's needs, guaranteeing the model performs reliably in real-world industrial scenarios.

\subsection{Mask Prediction and Angle-adaptive Cropping} \label{sec:angle}

To obtain the well-aligned cropped ACF code, we have to measure the oblique angle between the predicted mask regarding the code and the horizontal line, which can be calculated by:

\begin{equation}
    \alpha = \frac{1}{2} \arctan \left( \frac{2(B_{\text{2nd}}^{\text{mix}} - \mu_{x} B_{\text{1st}}^{y})}{B_{\text{2nd}}^{x} - B_{\text{2nd}}^{y} - \mu_{x} (B_{\text{1st}}^{x} - \mu_{x} B_{\text{1st}}^{x}) + \mu_{y} (B_{\text{1st}}^{y} - \mu_{y} B_{\text{1st}}^{y})} \right),
\end{equation}

where $B_{\text{1st}}^{x}$ and $B_{\text{1st}}^{y}$ are the first-order raw moments in the $x$ and $y$ directions, which represent the sums of the coordinates weighted by the pixel intensities in the mask. $B_{\text{2nd}}^{\text{mix}}$ is the mixed second-order raw moment, representing the sum of the product of the $x$ and $y$ coordinates weighted by the pixel intensities. $B_{\text{2nd}}^{x}$ and $B_{\text{2nd}}^{y}$ are the second-order raw moments in the $x$ and $y$ directions, representing the sums of the squares of the coordinates weighted by the pixel intensities. $\mu_{x}$ and $\mu_{y}$ are the centroid coordinates in the $x$ and $y$ directions. These are calculated as $B_{\text{1st}}^{x}$ divided by the summation across all of the $x$ and $y$ coordinates of the predicted 2D binary mask ($\sum mask[:,:]$) and $B_{\text{1st}}^{y}$ divided by the summation across all of the $x$ and $y$ coordinates of the predicted 2D binary mask ($\sum mask[:,:]$), respectively. This gives the average position of the mask in the directions for $x$ and $y$.

\subsection{Performance Metric} \label{sec:pm}

In this work, traditional segmentation performance metrics such as Intersection over Union (IoU) \cite{gilg2024we} or Dice Coefficient \cite{deng2024prpseg} could not be utilised due to the absence of ground truth masks for the datasets $\mathcal{D}_{t_2}$ and $\mathcal{D}_{t_3}$. Additionally, pre-trained SAM is unable to generate high-quality masks due to the existence of uneven surface textures, particularly those extending from the printed ACF code area regardless of its prompt type (point or box), which can be revealed in Fig. \ref{fig:overall_comparison}. Instead, we employed a proprietary screening tool provided by our industrial partner, which provides a justification of whether a cropped ACF code meets industrial regular standards or not. This metric considers factors such as alignment, completeness, and clarity of the cropped ACF code.

In addition to the adopted segmentation metric, due to the ultimate goal of the work is to feed the cropped image to a classifier to classify where the product is manufactured, factory $F_{1}$ or $F_{2}$ (see Fig. \ref{fig:ds}). To this end, we adopt a pre-trained ResNet34 \cite{he2015deep} model to produce the binary classification accuracy as the second performance metric. Such a dual-metric approach ensures that both segmentation and classification performances are optimised.


\subsection{Pseudo Re-adaptation} \label{sec:pr}

Pseudo Re-adaptation (PR), which is directly linked to the EAP (Sec. \ref{sec:eap}) and serves as a specific means to resolve the optimisation problem formulated in Eq. (\ref{dq:opt}), is a critical component of our AMRF. It is designed to refine augmentation operations that could maximise segmentation and cropping precision. After the initial training of the segmentation model on dataset $\mathcal{D}_{t_1}$ and subsequent testing on $\mathcal{D}_{t_2}$ and $\mathcal{D}_{t_3}$, we identify the images where the model's performance was sub-optimal. Concretely, 5\% of the images not segmented properly are selected for PR. These challenging images, with the process of PR, provide valuable insights into how the data should augmented to approach the model's perception range as expressed in our motivation visualised in Fig. \ref{fig:mot}(e).

Our hypothesis of newly involved augmentation methods is based on the assumption that if an image is inappropriately segmented and cropped, this image is either too sharp or too blurred, which prevents it from being efficiently sensed by the currently trained model. This, in turn, naturally inspired us to construct a pair of opposite augmentation methods -- Gaussian blur and contrast. To be concrete, if the image comes with high resolution and quality but cannot be sensed by the existing segmentation model, it indicates that the image could be blurred to a certain extent. Similarly, if the image is too blurred, image super-resolution is less preferred as an exponentially increased computation demand will be required while the risk of losing the essential features of specific granularity could occur. In this case, using an appropriate range of increasing or decreasing contrast will be a reasonable trade-off.

In this work, we adopted a pre-trained SAM for segmentation evaluations compared to the conventional models we use where the box prompts were provided. As such, different from the conventional model we are trying to fine-tune, the prompt-based SAM does not have any issue about whether the complete ACF code region is from a scale perspective. We decided to fine-tune the baseline segmentation model based on the observation of whether a sufficient scale range is given. Then, we tried to select from either or both of them to maximise the segmentation cropping as well as the classification performance with a minimum set of operations included in the pool. Such a problem can be empirically solved by fine-tuning the model in an accumulated effect.



\section{Experiments}


\subsection{Experimental Settings}

\subsubsection{Segmentation Baseline.} We adopted two conventional deep segmentation networks as baseline models for the proposed AMRF, \emph{i.e.} FCN (with a backbone of ResNet101 \cite{he2015deep})  and U-Net. Besides, we included a pre-trained SAM model with a Vision Transformer base and a patch size of $16 \times 16$ pixels (\emph{i.e.} ViT-B/16) \cite{dosovitskiy2021image} as one of three backbones, the others are ViT-L and ViT-H. To ensure fair comparisons among baseline models, we provide manually annotated boxes (purple dotted rectangles in Figs. \ref{fig:mot} and \ref{fig:framework}) as the prompts to the pre-trained SAM for zero-shot mask prediction.



\subsubsection{More implementation details.} We implemented AMRF on an NVIDIA RTX 3090 Ti GPU. To ensure uniform input image size of deep segmentation networks, we scale them to a resolution of $512 \times 512$ pixels with methods contained in the segmentation pool (depicted in Fig. \ref{fig:framework} and detailed in Sec. \ref{sec:eap}) applied. The pool $\mathcal{A}$ initially contains rotation in the range of $[-180^{\circ}, 180^{\circ}]$; brightness, saturation and contrast factors are drawn from the range of $[0.5, 1.5]$; and the hue factor is valued between -0.5 and 0.5. The Binary Cross Entropy loss \cite{nagendra2024patchrefinenet} is adopted for training and fine-tuning the two lightweight segmentation networks. For training, each baseline model was trained with a batch size of 16 for 200 epochs. In the case of fine-tuning the trained baseline model, 300 epochs are required for each of the newly added augmentation methods. The Adam optimiser is adopted with an initial learning rate of 0.001 and a weight decay of 0.00001.

\subsection{Optimal Scaling Range Selection} \label{sec:oss}

As the proposed ARMF was conducted in an accumulative manner, it is important to determine the zoom-in/out scaling range to ensure the default augmentation pool $\mathcal{A}$ is sufficiently optimised. To this end, the first part of the experiments began by adjusting the zoom-in/out factor $z$, starting with the default range of $[0.9, 1.1]$ as the baseline for FCN and U-Net. Though the cropping performance concerning the number of inappropriately cropped ACF codes produced by two factories $F_1$ and $F_2$ sharply exceeds the three pre-trained SAM as reported in Table \ref{tab:perf_1_deails}, FCN and U-Net were less competitive in $\mathcal{D}_{t_2}$ in binary classification performance reported by pre-trained ResNet34 model, lagged behind the most competitive SAM (ViT-L) by 2.19\% and 5.34\% respectively in Table \ref{tab:perf_1_deails}. Concretely, the default $z$ range could be flawed by observing the first, third and fourth row of the column `FCN (baseline)' in Fig. \ref{fig:fcn_detailed_compare} where the three failed croppings corresponded to the incompletely predicted mask, which was resulted by the insufficient zoom-in/out scale during the training phase.


\begin{table}[!htbp]
  \caption{Detailed comparisons of segmentation results on datasets $\mathcal{D}_{t_2}$ and $\mathcal{D}_{t_3}$ with respect to factories $F_1$ and $F_2$. The zoom-in/out, Gaussian blur and contrast factors are represented using $z$, $g$ and $c$. Light and dark grey rows correspond to the discussions in Secs. \ref{sec:oss} and \ref{sec:oed}.
  }
  \label{tab:perf_1_deails}
  \centering
  \scalebox{0.84}{
  \begin{tabular}{@{}lcccc@{}}
    \toprule
    Model &  $|\mathcal{D}^{F_1}_{t_2}|_{\text{fail/total}}$ &$|\mathcal{D}^{F_2}_{t_2}|_{\text{fail/total}}$ &  $|\mathcal{D}^{F_1}_{t_3}|_{\text{fail/total}}$ &$|\mathcal{D}^{F_2}_{t_3}|_{\text{fail/total}}$\\
    \midrule
    \rowcolor{Gray} SAM (ViT-B) & 117/1389 & 113/327 & 24/375 & 85/718 \\
    \rowcolor{Gray} SAM (ViT-L) & 95/1389 & 110/327 & 16/375 & 123/718 \\
    \rowcolor{Gray} SAM (ViT-H) & 102/1389 & 116/327 & 23/375 & 114/718 \\
    \midrule
    \rowcolor{Gray} FCN (baseline) & 46/1389 & 15/327 & 18/375 & 22/718 \\
    \rowcolor{Gray} FCN ($z \in [0.5, 2]$) & 5/1389 & 0/327 & 9/375 & 16/718\\
    \rowcolor{DarkGray} FCN ($z \in [0.5, 2]$, $g \in [1, 11]$) & 3/1389 & 0/327 & 7/375 & 15/718\\
    \rowcolor{DarkGray} FCN ($z \in [0.5, 2]$, $c \in [0.1, 1.5]$) & 1/1389 & 0/327 & 2/375 & 5/718\\
    \midrule
    \rowcolor{Gray} U-Net (baseline) & 93/1389 & 39/327 & 19/375 & 49/718 \\
    \rowcolor{Gray} U-Net ($z \in [0.5, 2]$) & 9/1389 & 2/327 & 8/375 & 32/718 \\
    \rowcolor{DarkGray} U-Net ($z \in [0.5, 2]$, $c \in [0.1, 1.5]$) & 5/1389 & 1/327 & 2/375 & 12/718 \\
  \bottomrule
  \end{tabular}
  }
\end{table}


On this basis, the zoom-in/out range $z$ was empirically determined to expand to $[0.5, 2]$ in accordance with pseudo re-adapation investigations performed on the two trained baseline models. Subsequently, the updated $z$ significantly expanded the perspective range of the fine-tuned model where cropping and classification accuracies were improved with a large margin. At this stage, the initial EAP $\mathcal{A}$ is optimised, which could be deemed as the proper timing to deal with more challenging factors entailed in the  ACF code collected in the open scenes.

\begin{table}[!htbp]
  \caption{Cropping and classification performance (in \%) comparisons among models trained on dataset $\mathcal{D}_{t_1}$ with evolving augmentation pool $\mathcal{A}$ with references to SAM with three different backbones. Colour codes are consistent with Table \ref{tab:perf_1_deails}.
  }
  \label{tab:perf_1}
  \centering
  \scalebox{0.9}{
  \begin{tabular}{@{}lcc@{}}
    \toprule
    Model & Acc$\big(\mathcal{M}_{\text{crop}}^{\mathcal{\theta(D}_{t_1})}(\mathcal{D}_{t_2}), \mathcal{M}^{\phi}_{\text{cla}}(\mathcal{D}_{t_2})\big)$ & Acc$\big(\mathcal{M}_{\text{crop}}^{\mathcal{\theta(D}_{t_1})}(\mathcal{D}_{t_3}), \mathcal{M}^{\phi}_{\text{cla}}(\mathcal{D}_{t_3})\big)$\\
    \midrule
    \rowcolor{Gray} SAM (ViT-B) & (86.60, 94.07) & (90.03, 81.75)\\
    \rowcolor{Gray} SAM (ViT-L) & (88.05, 94.13) & (87.28, 80.85)\\
    \rowcolor{Gray} SAM (ViT-H) & (87.30, 94.12) & (87.47, 81.66)\\
    \midrule
    \rowcolor{Gray} FCN (baseline) & (96.45, 91.94) & (96.34, 82.78)\\
    \rowcolor{Gray} FCN ($z \in [0.5, 2]$) & (99.71, 96.42) & (97.71, 86.07)\\
    \rowcolor{DarkGray} FCN ($z \in [0.5, 2]$, $g \in [1, 11]$) & (99.83, 96.53) & (97.99, 86.81)\\
    \rowcolor{DarkGray} FCN ($z \in [0.5, 2]$, $c \in [0.1, 1.5]$) & (99.94, 97.21) & (99.36, 86.82)\\
    \midrule
    \rowcolor{Gray} U-Net (baseline) & (92.31, 88.79) & (93.78, 80.81)\\
    \rowcolor{Gray} U-Net ($z \in [0.5, 2]$) & (99.36, 96.33) & (96.34, 85.94)\\
    \rowcolor{DarkGray} U-Net ($z \in [0.5, 2]$, $c \in [0.1, 1.5]$) & (99.65, 96.81) & (98.72, 86.33)\\
  \bottomrule
  \end{tabular}
  }
\end{table}


\subsection{Optimal EAP Determination} \label{sec:oed}

\begin{figure}[!ht]
  \centering
  \includegraphics[width=.95\linewidth]{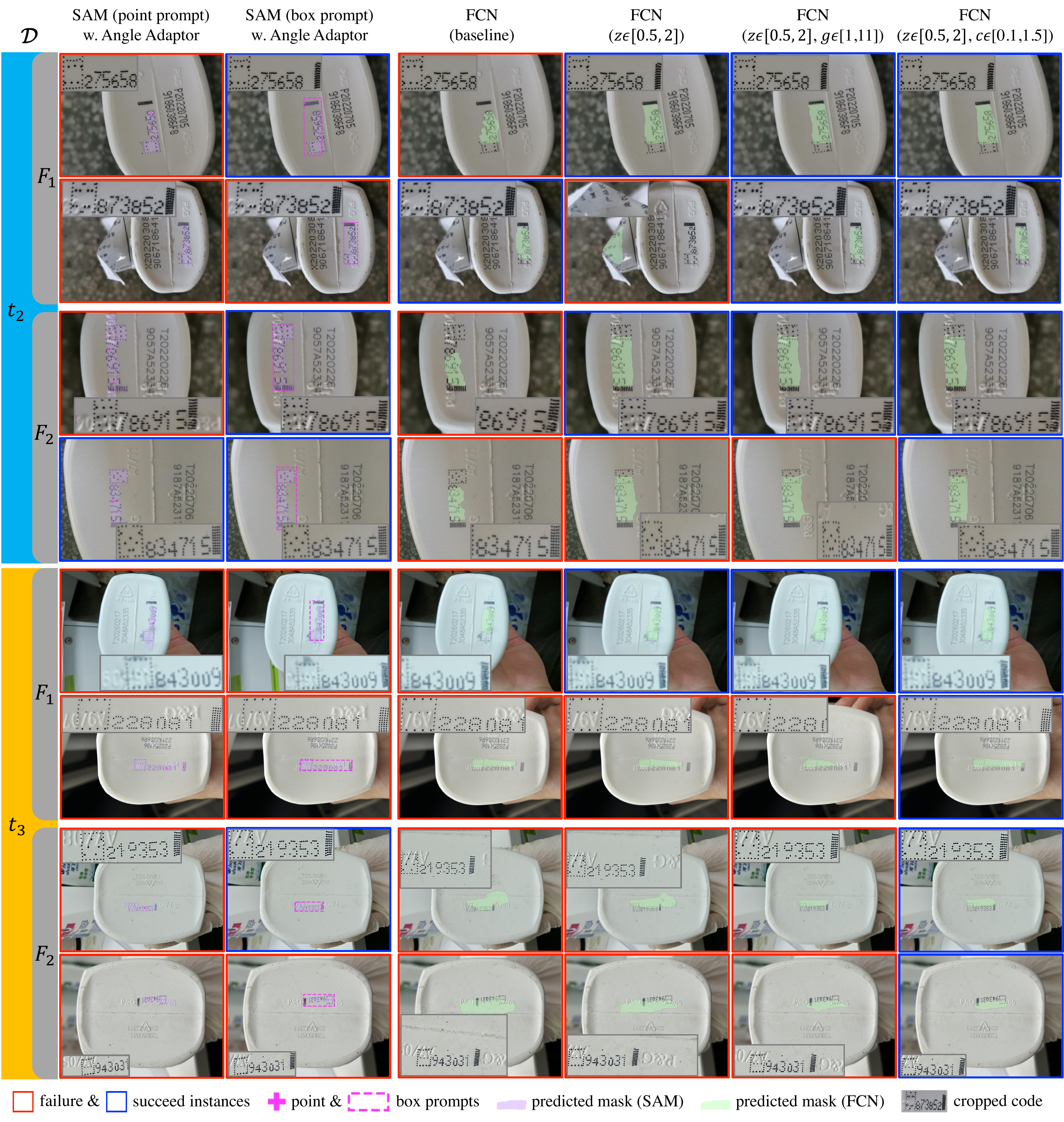}
  \caption{Detailed comparisons of segmentation and cropping among pre-trained SAM and FCN baseline where our proposed ARMF framework is applied.
  }
  \label{fig:fcn_detailed_compare}
\end{figure}

Though the competitive performance was achieved by fine-tuned model where $z \in [0.5, 2]$ with the initially optimised $\mathcal{A}$, we observed that uneven surface textures, particularly the ones emerging from the inside of the ACF and extending to the periphery, posed significant challenges to the trained model during testing (rows 4, 7 and 8 of the fourth column in Fig. \ref{fig:fcn_detailed_compare}). Besides, such difficulty was also evident in the pre-trained SAM (second column rows 2, 5, 6 and 8 of Fig. \ref{fig:fcn_detailed_compare}) when using manually given box prompts annotated using \textcolor{purple}{purple} dashed rectangle.

As the hypothesis explained in Sec. \ref{sec:pr} for introducing variations in image granularity, we experiment with two opposite augmentation techniques -- Gaussian blur and contrast adjustment -- in combination with the current $\mathcal{A}$ when fine-tuning the FCN for determining the optimal augmentation pool, whereas the U-net was not tuned with the former augmentation due to its drawback in learning global feature representations.

When fine-tuning Gaussian blur with the range of $[1,11$, the trained FCN has been proven to possess the ability to deal with several plastic carving textures that mixed with the printed ACF codes yet still failed in cases marked with \textcolor{red}{red} rectangles listed in the second to last column of Fig. \ref{fig:fcn_detailed_compare} where fine dust and stains can also trick the model into including them in the predicted mask. On the opposite, the increased or reduced contrast level applied in the fine-tuning process enable a more adaptable segmentation and cropping.

\begin{figure}[!htbp]
  \centering
  \includegraphics[width=.98\linewidth]{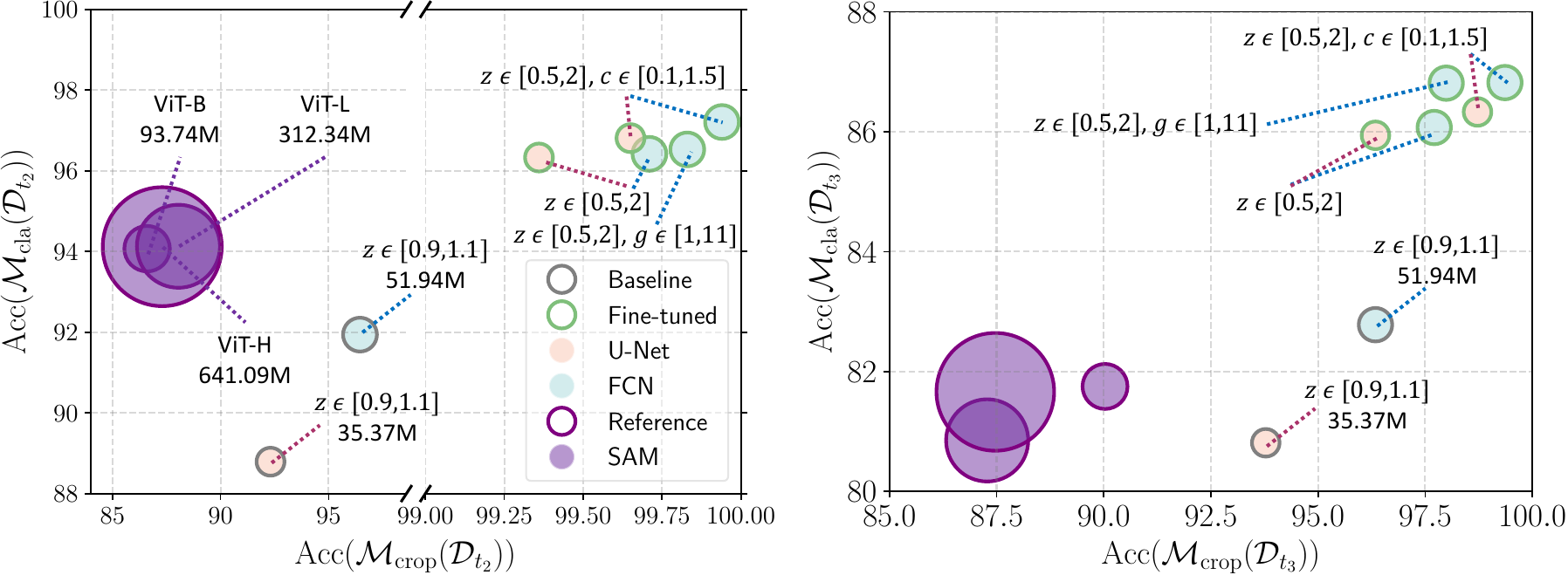}
  \caption{Cropping performance yielded by conventional (in conjunction with the proposed ARMF) and the latest prompt-based segmentation networks versus their associated classification of the cropped ACF codes using a pre-trained ResNet34 on temporally continuous datasets $\mathcal{D}_{t_2}$ and $\mathcal{D}_{t_3}$ introduced in Sec. \ref{sec:ds}. The radius of the bubbles denotes the number of trainable parameters in the model (U-Net: 35.37M, FCN: 51.94M, ViT-B: 93.74M, ViT-L: 312.34M, ViT-H: 641.09M). The colour denotes whether the model was a baseline model (grey), a model fine-tuned on our dataset (green) or a model given manually annotated references (purple).
  }
  \label{fig:overall_comparison}
\end{figure}

The Gaussian blur operation was excluded from the final EAP $\mathcal{A}$ due to its drawback of erasing out the discriminative features and introducing unpredictable bias for appropriate ACF code crop though a small amount of performance gain was obtained. Consequently, the final $\mathcal{A}$ is mainly constituted by zoom-in/out $z$ and contrast adjustment $c$ for both FCN and U-Net. The overall comparison among all the combinations of augmentation approaches with the references to pre-trained SAM are visualised in Fig.  \ref{fig:overall_comparison} where the trade-off between computational cost and ACF code cropping as well as the associated classification performance was achieved by the involvement of the proposed ARMF.


\section{Conclusion}

This paper presented AMRF for robust image segmentation in industrial inspection. Our approach leverages a dynamic and evolving pool of data augmentation techniques to address challenges posed by varying data qualities and small sample sizes, enhancing model generalisation and performance. The iterative refinement of the augmentation pool ensures the model remains robust and adaptable to new challenges, resulting in improved segmentation accuracy across diverse datasets.

Our experiments demonstrated significant performance gains with the AMRF, as fine-tuned FCN and U-Net models showed substantial improvements over baseline versions and transformer-based model SAM. Specifically, the fine-tuned FCN and U-Net models achieved notable increases in cropping and classification accuracies on challenging datasets. These results underscore the potential of our framework to enhance image segmentation models' performance and robustness in industrial applications, offering a promising solution for achieving high-precision segmentation with minimal training costs. An investigation of adopting the proposed ARMF to fine-tune SAM with a distillation of redundant weights could be an active future work.


\section*{Acknowledgment}
This work was supported by Procter \& Gamble (P\&G) United Kingdom Technical Centres Ltd.

%
%
\bibliographystyle{splncs04}
\bibliography{main}
\end{document}